\documentclass[
]{ceurart}

\usepackage{listings}
\usepackage{caption}
\usepackage{draftwatermark}
\SetWatermarkText{De-Factify 4.0 Workshop at AAAI 2025}
\SetWatermarkAngle{0}
\SetWatermarkScale{0.11}
\SetWatermarkVerCenter{0.95\paperheight}
\begin{document}

\copyrightyear{2025}
\copyrightclause{Copyright for this paper by its authors.
  Use permitted under Creative Commons License Attribution 4.0
  International (CC BY 4.0).}

\conference{De-Factify'25: 4\textsuperscript{th} Workshop on Multimodal Fact Checking and Hate Speech Detection,
Feb 25 -- Mar 4, 2025, Philadelphia, PA}

\title{SKDU at De-Factify 4.0: Natural Language Features for AI-Generated Text-Detection}


\author[1]{Shrikant Malviya}[%
orcid=0000-0002-7539-3721,
email=s.kant.malviya@gmail.com,
url=https://skmalviya.bitbucket.io/,
]
\cormark[1]
\address[1]{Department of Computer Science, Durham University, UK}

\author[2]{Pablo Arnau-González}[%
orcid=0000-0001-9048-4659,
email=pablo.arnau@uv.es,
]

\address[2]{Departament d'Informàtica, Universitat de València, Spain}

\author[2]{Miguel Arevalillo-Herráez}[%
orcid=0000-0002-0350-2079,
email=miguel.arevalillo@uv.es,
]

\author[1]{Stamos Katsigiannis}[%
orcid=0000-0001-9190-0941,
email=stamos.katsigiannis@durham.ac.uk,
]

\cortext[1]{Corresponding author.}

\begin{abstract}
The rapid advancement of large language models (LLMs) has introduced new challenges in distinguishing human-written text from AI-generated content. In this work, we explored a pipelined approach for AI-generated text detection that includes a feature extraction step (i.e. prompt-based rewriting features inspired by RAIDAR and content-based features derived from the NELA toolkit) followed by a classification module. Comprehensive experiments were conducted on the Defactify4.0 dataset, evaluating two tasks: binary classification to differentiate human-written and AI-generated text, and multi-class classification to identify the specific generative model used to generate the input text. Our findings reveal that NELA features significantly outperform RAIDAR features in both tasks, demonstrating their ability to capture nuanced linguistic, stylistic, and content-based differences. Combining RAIDAR and NELA features provided minimal improvement, highlighting the redundancy introduced by less discriminative features. Among the classifiers tested, XGBoost emerged as the most effective, leveraging the rich feature sets to achieve high accuracy and generalisation.
\end{abstract}

\begin{keywords}
  AI-Generated Text Detection \sep
  Natural Language Features \sep
  Rewriting features \sep
  Text Classification
\end{keywords}

\maketitle

\vspace{-0.6cm}
\section{Introduction}
The rapid advancements in large language models (LLMs) have revolutionised natural language processing (NLP), enabling systems to generate human-like text with remarkable fluency and contextual relevance. These models, such as OpenAI's GPT-3.5 \cite{brown2020language}, GPT-4 \cite{openai2024gpt4}, and Meta's LLaMA \cite{touvron2023llama}, are now widely used across various domains, including education, customer support, and creative content generation. While these systems have demonstrated significant potential to augment human productivity and assist in a variety of tasks, they also introduce critical challenges, particularly in ensuring their responsible use.


One major concern is the misuse of LLMs for generating misleading or harmful content \cite{elsafoury2024bias}. Malicious actors can exploit these models to create misinformation, spam, or fraudulent websites, leveraging their ability to produce seemingly credible and factually sound text at scale \cite{bender2021dangers}. Compounding the issue, these models often present information with high fluency and confidence, which can deceive users into mistaking coherence for truthfulness \cite{rise,junk}. Inexperienced users may unknowingly rely on AI-generated text for high-stakes tasks, such as academic writing or medical advice, only to encounter significant inaccuracies or unintended consequences. This raises concerns about the ethical and safe deployment of these powerful systems in everyday applications \cite{weidinger2021ethical}.

Given the proliferation of AI-generated content, it is essential to develop a text detector that not only processes a piece of text but also identifies its source, as the ability to distinguish between human-written and AI-generated text has become a pressing need. This task, however, is increasingly challenging as the sophistication of LLMs continues to evolve \cite{sadasivan2024can}. Traditional methods, including statistical measures, feature-based classifiers, and fine-tuned language models, have shown varying levels of success, but face limitations in generalisability and robustness \cite{kumarage2024survey}. For example, current detection systems often struggle to adapt to novel architectures or combat adversarial tactics such as paraphrasing \cite{mao2023raidar}. Furthermore, the lack of comprehensive benchmarks that incorporate stylistic, structural, and sequence-based features limits progress in developing effective detection mechanisms~\cite{horne2018assessing,opara2024styloai}.

In this work, we investigate the task of identifying AI-generated text, focusing on evaluating features such as geneRative AI Detection viA Rewriting (RAIDAR) \cite{mao2023raidar}, content-based features from the News Landscape (NELA) toolkit \cite{horne2017just}, e.g.,
stylistic features (stopwords, punctuation, quotes, negations etc), complexity features (type-token ratio, words per sentence, noun/verb phrase syntax tree depth), psychology features (linguistic inquiry and word count \cite{tausczik2010psychological},  positive and negative emotions \cite{thelwall2010sentiment}). Specifically, we explore their performance in distinguishing between human-written and AI-generated text, supplemented by an analysis of stylistic, structural, and sequence-based features. The findings of this research provide critical insights into the differences between human- and AI-generated text, offering valuable guidance for improving detection frameworks. By establishing benchmarks and highlighting the limitations of existing methods, this work contributes to the development of more robust and adaptable tools for mitigating the risks associated with AI-generated content. Ultimately, our study seeks to foster the responsible use of LLMs and support the creation of safer digital environments. We make our code publicly available \url{ https://github.com/skmalviya/ai-gen\_text\_defactify}.

In short, the contributions of this paper are the following:
\vspace{-0.3cm}
\begin{itemize}
    \item[-] We demonstrate a pipelined approach for AI-generated text detection, integrating feature extraction followed by a classification module.
    \item[-] We provide a detailed comparative analysis of rewriting-based and content-based feature extraction approaches, highlighting the strengths and limitations of each.
    \item[-] Through comprehensive experiments, we demonstrate that content-based features significantly outperform rewriting-based features in both binary and multi-class classification tasks.
    \item [-] We evaluate the effectiveness of popular classifiers, identifying XGBoost as the most suitable model for handling diverse feature sets in AI-generated text detection.
\end{itemize}

\section{Shared Task CT\textsuperscript{2}: AI-Generated Text Detection} \label{method}
\subsection{Tasks}
The shared task \textit{CT\textsuperscript{2}: AI-Generated Text Detection} challenge is an initiative organised by researchers from the University of South Carolina, US, to address the growing concerns surrounding AI-generated text. With the increasing sophistication of LLMs such as GPT-4, LLaMA, Qwen, Gemma, Mistral, and Yi, distinguishing human-authored text from AI-generated text has become a critical task in fields like media authenticity, misinformation detection, and AI ethics. The challenge provides participants with a curated dataset and a platform to develop and evaluate machine learning models aimed at identifying and attributing AI-generated text.

To address these challenges, they propose an \textit{AI-Generated Text Detection} shared task \cite{roy2025defactify_overview_text}, accompanied by a comprehensive dataset designed to support two primary classification tasks: (a) \textbf{Task A}: Binary classification to determine whether a given text is human-authored or AI-generated. (b) \textbf{Task B}: Multi-class classification to identify the specific AI model responsible for generating synthetic text.
The dataset and tasks aim to enable the development of robust machine learning models capable of accurately classifying and attributing text generation, thereby advancing the field of generative AI forensics.
\vspace{-0.3cm}
\subsection{Dataset Description}
The dataset \cite{roy2025defactify_dataset_text} for the task consists of textual data (train: $50k$, validation: $10k$ and test: $10k$, approximately) annotated with two types of labels: (a) \textbf{Label\_A}: A binary label indicating whether the text is human-authored (\texttt{0}) or AI-generated (\texttt{1}). (b) \textbf{Label\_B}: A multi-class label that identifies the specific AI model responsible for generating the text. The available models include: (1) Gemma-2-9B, (2) Mistral-7B, (3) Qwen-2-72B, (4) LLaMA-8B, (5) Yi-Large, and (6) GPT-4-o. Additionally, each row in the dataset includes a unique index for identification and the corresponding text entry. Human-generated text entries are labeled as \texttt{0} for \textbf{Label\_A}, with \textbf{Label\_B} remaining a placeholder (\texttt{Human\_Story} or similar).

The dataset is structured to support seamless data ingestion and preprocessing for machine learning tasks. Each entry includes the following fields: (i) \textbf{Index}: A unique identifier for each text instance. (ii) \textbf{Text}: The text content, either human-authored or AI-generated. (iii) \textbf{Label\_A}: Binary label for Task A. (iv) \textbf{Label\_B}: Multi-class label for Task B.
Examples from the development dataset are shown in Table~\ref{tab:examples}.

\begin{table}[t]
    \centering
    \caption{Example rows from the dataset.}
    \label{tab:examples}
    \resizebox{0.7\columnwidth}{!}{
    \begin{tabular}{|c|p{8cm}|c|c|}
        \hline
        \textbf{Index} & \textbf{Text} & \textbf{Label\_A} & \textbf{Label\_B} \\ \hline
        0 & The U.S. bombings that ended World War II didn’t mark the close of atomic... & 0 & Human\_Story \\ \hline
        1 & The radioactive fallout has led to a spike in thyroid cancers, leukemia... & 1 & Qwen-2-72B \\ \hline
        2 & Climate models predict a 50\% increase in extreme weather events by 2050... & 1 & GPT-4-o \\ \hline
    \end{tabular}
    }
\end{table}

\vspace{-0.3cm}
\section{Methodology}
\vspace{-0.3cm}
In this section, we present our methodology for tackling the AI-generated text detection challenge. Our approach leverages a combination of feature extraction techniques, inspired by state-of-the-art methods, to maximise detection accuracy. Specifically, we employ two complementary feature extraction strategies: (1) leveraging rewriting-based feature extraction, inspired by RAIDAR: Generative AI Detection via Rewriting, and (2) incorporating NELA features from a Multi-Module Toolkit named `NELA'. These features are integrated into a unified pipeline designed for efficient training and testing.

\subsection{Our Pipeline}
\begin{figure}[!htb]
  \centering
  \includegraphics[width=0.9\textwidth]{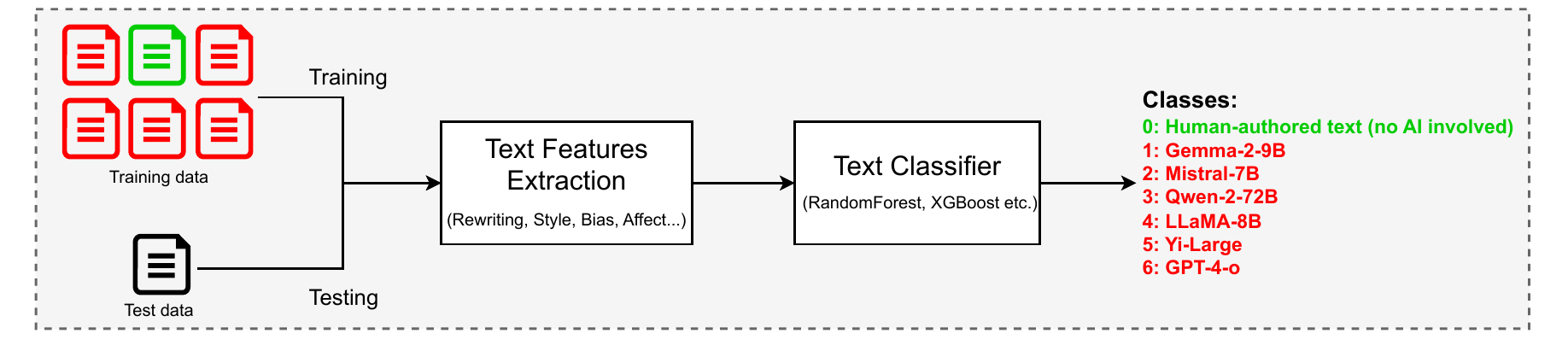}
  \caption{Overview of the Training/Testing pipeline for text classification tasks. The pipeline includes text preprocessing, feature extraction (rewriting-based/NELA features), feature fusion, classifier training, and evaluation.}
  \label{img:proposed_pipeline}
\end{figure}

Our methodology follows a systematic pipeline for training and testing, as depicted in Figure~\ref{img:proposed_pipeline}. The pipeline ensures a clear flow from input text to final predictions, with well-defined stages for feature extraction, fusion, and classification.

\vspace{-0.8cm}
\subsection{Feature Extraction}

Our feature extraction approach combines rewriting-based techniques and NELA features to capture diverse linguistic and stylistic attributes. Inspired by the RAIDAR framework \cite{mao2023raidar}, we adopt a rewriting approach to detect AI-generated text. The underlying intuition is that human-written and AI-generated text respond differently to perturbations in the form of rewriting. This step highlights structural and stylistic differences based on rewriting features. In parallel, NELA features \cite{horne2018assessing} contribute lexical, stylometric, and content-based attributes, such as word frequencies, readability scores, sentiment analysis, and entity recognition. By integrating these two methodologies, we ensure the extraction of both fine-grained and broad textual features, enhancing the ability to distinguish between human-written and AI-generated text.
\begin{itemize}
    \item \textit{Prompt-based Rewriting Features (RAIDAR)}: For the rewriting-based feature extraction, we utilised the RAIDAR-inspired module, which performs paraphrasing and syntactic transformations on the input text \cite{mao2023raidar}. The rewriting for a given text is done based on 7 different prompts (Appendix~\ref{appx:prompts}). All prompts, including the input text, are considered for feature extraction. For prompting, we use \texttt{meta-llama/Llama-3.1-8B}\footnote{\url{https://huggingface.co/meta-llama/Llama-3.1-8B}} model for generating rewritings due to its balance between computational efficiency and generative capabilities. Compared to larger models, it provided sufficient quality in text rewriting while maintaining reasonable inference costs. Later, RAIDAR extracts features highlighting statistics of common terms, fuzzy ratios, etc., for each example.
    \item \textit{Content-based Features (NELA)}: The NELA feature extraction module uses the open-source NELA-Toolkit to extract content-based attributes of three broad categories: stylistic, complexity, and psychological \cite{horne2017just}. Lexical features include word frequencies and sentiment scores, while stylistic features capture readability, punctuation patterns, and average word length; psychological features, on the other hand, include sentiment analysis and the use of the Linguistic Inquiry and Word Count (LIWC) dictionaries~\cite{tausczik2010psychological}. For a given text, it extracts 87 attributes across all three categories.
\end{itemize}

\subsection{Training}

The training stage aims to build a robust classifier for distinguishing human-written text from AI-generated content and identifying specific generative models. The process begins with extracting features by prompt-based rewriting (RAIDAR) and the NELA toolkit. We conducted experiments to evaluate the strength of each feature set individually and also combined them into a unified representation. Each feature set is used for training machine learning classifiers, such as Random Forests, XGBoost, and Support Vector Machines, due to their effectiveness in handling structured feature-based text classification tasks \cite{oktafia2022comparison,umam2024performance}. Each classifier is trained with their default parameter setting. e.g. (scikit-learn v1.5.2).

\subsection{Testing}

The testing stage evaluates the performance of the trained classifier on unseen data. Similar feature-extraction steps are applied to the test input to ensure compatibility with the training data. Features are extracted using the same rewriting (RAIDAR) and NELA modules, ensuring consistency in representation. The classifier’s predictions are compared against ground truth labels to compute the evaluation metric F1-score. This stage highlights the generalisation ability of the model and its effectiveness in handling real-world data.

\section{Results \& Discussion}

To evaluate our approach, we conducted experiments on the AI-generated text detection dataset, focusing on the two tasks of the challenge: Task A, involving binary classification to distinguish human-written text from AI-generated content, and Task B, requiring multi-class classification to identify the specific generative model. Various classifiers, including Support Vector Machines (SVC), Random Forests, and XGBoost, were evaluated on the development dataset. XGBoost emerged as the best-performing classifier due to its ability to handle complex feature sets and its robust performance across tasks. This classifier was subsequently used for testing on unseen data to assess generalisation capabilities. The F1 score metric is used to evaluate the pipeline's performance comprehensively.

On the development dataset, we evaluated the performance of various classification methods, such as SVC, Random Forest, and XGBoost. Table~\ref{tab:dev_result} shows XGBoost consistently outperformed SVC and Random Forest across all combinations of feature sets, achieving an F1-score (Defactify4.0 challenge score) of 0.9979 for Task A and 0.8489 for Task B. Its ability to model complex, nonlinear relationships between features allowed it to better exploit the rich information embedded in NELA features. SVC struggled with the high-dimensional feature space, while Random Forest exhibited reduced performance due to its tendency to overfit on redundant features from the RAIDAR module. Using XGBoost, we obtained an F1-score score of 0.9945 and 0.7615 on the testing leaderboard for Task A and B, respectively (Table~\ref{tab:test_result}).

\begin{table*}[t]
  \caption{Performance on the development set.}
  \label{tab:dev_result}
  \resizebox{0.9\columnwidth}{!}{
  \begin{tabular}{lcccccc}
    \toprule

\multirow{2}{*}{Feature-Set} & \multicolumn{2}{c}{ SVC } & \multicolumn{2}{c}{ Random Forest } & \multicolumn{2}{c}{ XGBoost } \\
\cmidrule(lr){2-3} 
\cmidrule(lr){4-5}
\cmidrule(lr){6-7}
    
     & F1 (Task-A) & F1 (Task-B) & F1 (Task-A) & F1 (Task-B) & F1 (Task-A) & F1 (Task-B)\\
    \midrule
    RAIDAR \cite{mao2023raidar} & 0.9573 & 0.5403 & 0.9548 & 0.5283 & 0.9652 & 0.5719\\
    NELA \cite{horne2017just} & 0.9205 & 0.4585 & 0.9942 & 0.8061 & \textbf{0.9979} & \textbf{0.8489}\\
    RAIDAR + NELA & 0.9268 & 0.4337 & 0.9933 & 0.7754 & 0.9968 & 0.8430 \\
  \bottomrule
\end{tabular}
}
\end{table*}

\begin{wraptable}{rh}{6cm}
\vspace{-0.8cm}
  \caption{ \small Performance on the test set.}
  \label{tab:test_result}
  \resizebox{0.38\columnwidth}{!}{
  \begin{tabular}{lcc}
    \toprule

\multirow{2}{*}{Feature-Set} & \multicolumn{2}{c}{ XGBoost } \\
\cmidrule(lr){2-3} 
    
     & F1 (Task-A) & F1 (Task-B) \\
    \midrule
    RAIDAR \cite{mao2023raidar} & 0.9454 & 0.4410
 \\
    NELA \cite{horne2017just} & \textbf{0.9945} & \textbf{0.7615} \\
    RAIDAR + NELA & 0.9917 & 0.7443 \\
  \bottomrule
\end{tabular}
}
\end{wraptable}

As shown in Tables~\ref{tab:dev_result} and~\ref{tab:test_result}, NELA features significantly outperformed RAIDAR-based features across both tasks on both the development and testing sets. With XGBoost, NELA features show a jump of 27\% and 32\% on F1 score compared to RAIDAR on Task-B for development and test data, respectively. However, combining RAIDAR and NELA features does not demonstrate any significant improvement in the performance on either Task A or Task B. These results suggest that RAIDAR features introduced redundancy rather than complementarity, likely due to their limited discriminatory capacity in the presence of high-quality NELA features. NELA features capture richer and more discriminative features for both the binary and multi-class classification task.

\begin{wrapfigure}{rh}{6cm}
  \centering
  \vspace{-0.2cm}
  \includegraphics[width=0.4\textwidth]{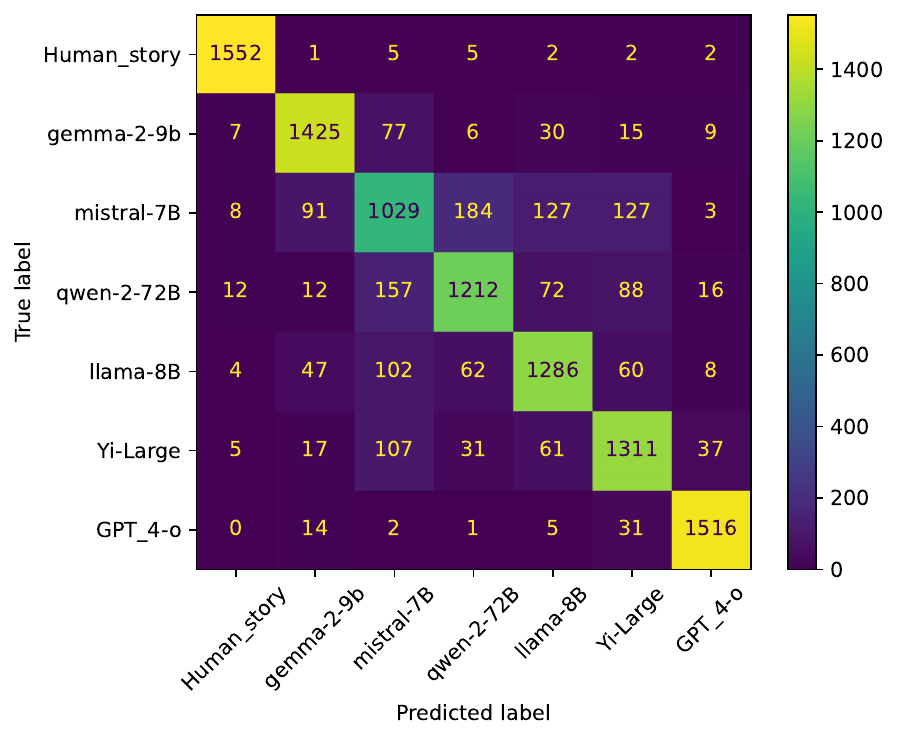}
  \caption{\footnotesize Confusion matrix illustrating the classification performance across human-written and AI-generated text classes.}
  \label{img:conf_matrix_results}
\end{wrapfigure}

To better understand the limitations of our detection approach, we analysed the misclassification patterns using a confusion matrix on the development data as in Figure~\ref{img:conf_matrix_results}. The classifier demonstrated strong performance in distinguishing human-written and GPT-4-o text, with 1552 and 1516, respectively, correctly classified instances. This suggests their consistent generative characteristics,  making it easier to classify them correctly. Gemma-2-9B and Qwen-2-72B showed notable overlap. There were 91 instances where Qwen-generated text was misclassified as Gemma-generated and 77 where Gemma was misclassified as Qwen. This could be due to similar training data distributions or shared linguistic patterns among these models. For Mid-Sized Models (LLaMA, Yi, Qwen),  LLaMA-8B and Yi-Large exhibited moderate misclassification rates. For instance, 88 instances of Yi-Large were misclassified as LLaMA-8B, and 60 cases of LLaMA-8B were mistaken for Yi-Large.

Overall, our findings highlight the importance of selecting features for AI-generated text detection, avoiding redundancy and ensuring effective integration. The superior performance of NELA features emphasises the need for contextually rich and stylistically diverse attributes in detecting AI-generated text. Despite RAIDAR’s theoretical appeal, its practical limitations in multi-class classification underscore the need for more robust rewriting-based approaches in future research.

To improve generalisability, future work will explore domain adaptation techniques, augmenting the training data with more diverse AI-generated text sources. Additionally, meta-learning strategies can be employed to finetune detection models across different LLM architectures. Investigating feature representations that remain invariant across various models could also enhance robustness.

\section{Conclusion}

This work addresses the challenge of AI-generated text detection by leveraging features such as prompt-based rewriting (RAIDAR) and content-based features (NELA). Through rigorous evaluation, we identified XGBoost as the optimal classifier for this task, due to its ability to exploit rich feature representations, and NELA features as the superior choice. Our results highlight the importance of selecting high-quality, discriminative features and avoiding redundancy in detection frameworks. This research contributes to the development of more robust tools for distinguishing AI-generated text, supporting ethical and responsible AI deployment. Future work could focus on improving rewriting-based feature extraction methods to enhance their complementarity with content-based approaches, further advancing the field of generative AI forensics.

\section*{Acknowledgments}
The authors in this project have been funded by UK EPSRC grant ``AGENCY: Assuring Citizen Agency in a World with Complex Online Harms'' under grant EP/W032481/2.  By Generalitat Valenciana under Grant CIAPOS/2022/163. and by project PID2023-150960NB-I00, funded by the Spanish Ministry of Science, Innovation and Universities and the European Union. 

This work used the DiRAC@Durham facility managed by the Institute for Computational Cosmology on behalf of the STFC DiRAC HPC Facility (www.dirac.ac.uk). The equipment was funded by BEIS capital funding via STFC capital grants ST/P002293/1, ST/R002371/1 and ST/S002502/1, Durham University and STFC operations grant ST/R000832/1. DiRAC is part of the National e-Infrastructure.
The authors gratefully acknowledge the computer resources at Artemisa and the technical support provided by the Instituto de Fisica Corpuscular, IFIC(CSIC-UV). Artemisa is co-funded by the European Union through the 2014-2020 ERDF Operative Programme of Comunitat Valenciana, project IDIFEDER/2018/048.
\appendix

\section{Rewriting prompts}
\label{appx:prompts}

The 7 prompts in the RAIDAR-based feature extraction approach were designed to elicit variations in text generation, emphasizing differences in syntactic structure, coherence, fluency, and factual consistency. These prompts were incorporated from prior research \cite{mao2023raidar} that demonstrated how different types of paraphrasing affect text features, making them valuable indicators for distinguishing AI-generated text from human-written content.

\begin{itemize}
  \item CONCISE: Concise this for me and keep all the information.
  \item REVISE: Revise this with your best effort
  \item POLISH: Help me polish this
  \item REWRITE: Rewrite this for me 
  \item FLUENT: Make this fluent while doing minimal change
  \item REFINE: Refine this for me please
  \item GPTIZE: Improve this in GPT way
\end{itemize}

\bibliography{sample-ceur}

\begin{thebibliography}{20}
\expandafter\ifx\csname natexlab\endcsname\relax\def\natexlab#1{#1}\fi
\providecommand{\url}[1]{\texttt{#1}}
\providecommand{\href}[2]{#2}
\providecommand{\path}[1]{#1}
\providecommand{\DOIprefix}{doi:}
\providecommand{\ArXivprefix}{arXiv:}
\providecommand{\URLprefix}{URL: }
\providecommand{\Pubmedprefix}{pmid:}
\providecommand{\doi}[1]{\href{http://dx.doi.org/#1}{\path{#1}}}
\providecommand{\Pubmed}[1]{\href{pmid:#1}{\path{#1}}}
\providecommand{\bibinfo}[2]{#2}
\ifx\xfnm\relax \def\xfnm[#1]{\unskip,\space#1}\fi
\bibitem[{Brown et~al.(2020)Brown, Mann, Ryder, Subbiah, Kaplan, Dhariwal, Neelakantan, Shyam, Sastry, Askell, Agarwal, {Herbert-Voss}, Krueger, Henighan, Child, Ramesh, Ziegler, Wu, Winter, Hesse, Chen, Sigler, Litwin, Gray, Chess, Clark, Berner, McCandlish, Radford, Sutskever, and Amodei}]{brown2020language}
\bibinfo{author}{T.~Brown}, \bibinfo{author}{B.~Mann}, \bibinfo{author}{N.~Ryder}, \bibinfo{author}{M.~Subbiah}, \bibinfo{author}{J.~D. Kaplan}, \bibinfo{author}{P.~Dhariwal}, \bibinfo{author}{A.~Neelakantan}, \bibinfo{author}{P.~Shyam}, \bibinfo{author}{G.~Sastry}, \bibinfo{author}{A.~Askell}, \bibinfo{author}{S.~Agarwal}, \bibinfo{author}{A.~{Herbert-Voss}}, \bibinfo{author}{G.~Krueger}, \bibinfo{author}{T.~Henighan}, \bibinfo{author}{R.~Child}, \bibinfo{author}{A.~Ramesh}, \bibinfo{author}{D.~Ziegler}, \bibinfo{author}{J.~Wu}, \bibinfo{author}{C.~Winter}, \bibinfo{author}{C.~Hesse}, \bibinfo{author}{M.~Chen}, \bibinfo{author}{E.~Sigler}, \bibinfo{author}{M.~Litwin}, \bibinfo{author}{S.~Gray}, \bibinfo{author}{B.~Chess}, \bibinfo{author}{J.~Clark}, \bibinfo{author}{C.~Berner}, \bibinfo{author}{S.~McCandlish}, \bibinfo{author}{A.~Radford}, \bibinfo{author}{I.~Sutskever}, \bibinfo{author}{D.~Amodei},
\newblock \bibinfo{title}{Language {{Models}} are {{Few-Shot Learners}}},
\newblock in: \bibinfo{booktitle}{Advances in {{Neural Information Processing Systems}}}, volume~\bibinfo{volume}{33}, \bibinfo{publisher}{Curran Associates, Inc.}, \bibinfo{year}{2020}, pp. \bibinfo{pages}{1877--1901}.
\bibitem[{OpenAI et~al.(2024)OpenAI, Achiam, Adler, Agarwal, Ahmad, Akkaya, Aleman, Almeida, Altenschmidt, Altman, Anadkat, Avila, Babuschkin, Balaji, Balcom, Baltescu, Bao, Bavarian, Belgum, Bello, Berdine, {Bernadett-Shapiro}, Berner, Bogdonoff, Boiko, Boyd, Brakman, Brockman, Brooks, Brundage, Button, Cai, Campbell, Cann, Carey, Carlson, Carmichael, Chan, Chang, Chantzis, Chen, Chen, Chen, Chen, Chen, Chess, Cho, Chu, Chung, Cummings, Currier, Dai, Decareaux, Degry, Deutsch, Deville, Dhar, Dohan, Dowling, Dunning, Ecoffet, Eleti, Eloundou, Farhi, Fedus, Felix, Fishman, Forte, Fulford, Gao, Georges, Gibson, Goel, Gogineni, Goh, {Gontijo-Lopes}, Gordon, Grafstein, Gray, Greene, Gross, Gu, Guo, Hallacy, Han, Harris, He, Heaton, Heidecke, Hesse, Hickey, Hickey, Hoeschele, Houghton, Hsu, Hu, Hu, Huizinga, Jain, Jain, Jang, Jiang, Jiang, Jin, Jin, Jomoto, Jonn, Jun, Kaftan, Kaiser, Kamali, Kanitscheider, Keskar, Khan, Kilpatrick, Kim, Kim, Kim, Kirchner, Kiros, Knight, Kokotajlo, Kondraciuk, Kondrich,
  Konstantinidis, Kosic, Krueger, Kuo, Lampe, Lan, Lee, Leike, Leung, Levy, Li, Lim, Lin, Lin, Litwin, Lopez, Lowe, Lue, Makanju, Malfacini, Manning, Markov, Markovski, Martin, Mayer, Mayne, McGrew, McKinney, McLeavey, McMillan, McNeil, Medina, Mehta, Menick, Metz, Mishchenko, Mishkin, Monaco, Morikawa, Mossing, Mu, Murati, Murk, M{\'e}ly, Nair, Nakano, Nayak, Neelakantan, Ngo, Noh, Ouyang, O'Keefe, Pachocki, Paino, Palermo, Pantuliano, Parascandolo, Parish, Parparita, Passos, Pavlov, Peng, Perelman, Peres, Petrov, Pinto, Michael, Pokorny, Pokrass, Pong, Powell, Power, Power, Proehl, Puri, Radford, Rae, Ramesh, Raymond, Real, Rimbach, Ross, Rotsted, Roussez, Ryder, Saltarelli, Sanders, Santurkar, Sastry, Schmidt, Schnurr, Schulman, Selsam, Sheppard, Sherbakov, Shieh, Shoker, Shyam, Sidor, Sigler, Simens, Sitkin, Slama, Sohl, Sokolowsky, Song, Staudacher, Such, Summers, Sutskever, Tang, Tezak, Thompson, Tillet, Tootoonchian, Tseng, Tuggle, Turley, Tworek, Uribe, Vallone, Vijayvergiya, Voss, Wainwright, Wang,
  Wang, Wang, Ward, Wei, Weinmann, Welihinda, Welinder, Weng, Weng, Wiethoff, Willner, Winter, Wolrich, Wong, Workman, Wu, Wu, Wu, Xiao, Xu, Yoo, Yu, Yuan, Zaremba, Zellers, Zhang, Zhang, Zhao, Zheng, Zhuang, Zhuk, and Zoph}]{openai2024gpt4}
\bibinfo{author}{OpenAI}, \bibinfo{author}{J.~Achiam}, \bibinfo{author}{S.~Adler}, \bibinfo{author}{S.~Agarwal}, \bibinfo{author}{L.~Ahmad}, \bibinfo{author}{I.~Akkaya}, \bibinfo{author}{F.~L. Aleman}, \bibinfo{author}{D.~Almeida}, \bibinfo{author}{J.~Altenschmidt}, \bibinfo{author}{S.~Altman}, \bibinfo{author}{S.~Anadkat}, \bibinfo{author}{R.~Avila}, \bibinfo{author}{I.~Babuschkin}, \bibinfo{author}{S.~Balaji}, \bibinfo{author}{V.~Balcom}, \bibinfo{author}{P.~Baltescu}, \bibinfo{author}{H.~Bao}, \bibinfo{author}{M.~Bavarian}, \bibinfo{author}{J.~Belgum}, \bibinfo{author}{I.~Bello}, \bibinfo{author}{J.~Berdine}, \bibinfo{author}{G.~{Bernadett-Shapiro}}, \bibinfo{author}{C.~Berner}, \bibinfo{author}{L.~Bogdonoff}, \bibinfo{author}{O.~Boiko}, \bibinfo{author}{M.~Boyd}, \bibinfo{author}{A.-L. Brakman}, \bibinfo{author}{G.~Brockman}, \bibinfo{author}{T.~Brooks}, \bibinfo{author}{M.~Brundage}, \bibinfo{author}{K.~Button}, \bibinfo{author}{T.~Cai}, \bibinfo{author}{R.~Campbell}, \bibinfo{author}{A.~Cann},
  \bibinfo{author}{B.~Carey}, \bibinfo{author}{C.~Carlson}, \bibinfo{author}{R.~Carmichael}, \bibinfo{author}{B.~Chan}, \bibinfo{author}{C.~Chang}, \bibinfo{author}{F.~Chantzis}, \bibinfo{author}{D.~Chen}, \bibinfo{author}{S.~Chen}, \bibinfo{author}{R.~Chen}, \bibinfo{author}{J.~Chen}, \bibinfo{author}{M.~Chen}, \bibinfo{author}{B.~Chess}, \bibinfo{author}{C.~Cho}, \bibinfo{author}{C.~Chu}, \bibinfo{author}{H.~W. Chung}, \bibinfo{author}{D.~Cummings}, \bibinfo{author}{J.~Currier}, \bibinfo{author}{Y.~Dai}, \bibinfo{author}{C.~Decareaux}, \bibinfo{author}{T.~Degry}, \bibinfo{author}{N.~Deutsch}, \bibinfo{author}{D.~Deville}, \bibinfo{author}{A.~Dhar}, \bibinfo{author}{D.~Dohan}, \bibinfo{author}{S.~Dowling}, \bibinfo{author}{S.~Dunning}, \bibinfo{author}{A.~Ecoffet}, \bibinfo{author}{A.~Eleti}, \bibinfo{author}{T.~Eloundou}, \bibinfo{author}{D.~Farhi}, \bibinfo{author}{L.~Fedus}, \bibinfo{author}{N.~Felix}, \bibinfo{author}{S.~P. Fishman}, \bibinfo{author}{J.~Forte}, \bibinfo{author}{I.~Fulford},
  \bibinfo{author}{L.~Gao}, \bibinfo{author}{E.~Georges}, \bibinfo{author}{C.~Gibson}, \bibinfo{author}{V.~Goel}, \bibinfo{author}{T.~Gogineni}, \bibinfo{author}{G.~Goh}, \bibinfo{author}{R.~{Gontijo-Lopes}}, \bibinfo{author}{J.~Gordon}, \bibinfo{author}{M.~Grafstein}, \bibinfo{author}{S.~Gray}, \bibinfo{author}{R.~Greene}, \bibinfo{author}{J.~Gross}, \bibinfo{author}{S.~S. Gu}, \bibinfo{author}{Y.~Guo}, \bibinfo{author}{C.~Hallacy}, \bibinfo{author}{J.~Han}, \bibinfo{author}{J.~Harris}, \bibinfo{author}{Y.~He}, \bibinfo{author}{M.~Heaton}, \bibinfo{author}{J.~Heidecke}, \bibinfo{author}{C.~Hesse}, \bibinfo{author}{A.~Hickey}, \bibinfo{author}{W.~Hickey}, \bibinfo{author}{P.~Hoeschele}, \bibinfo{author}{B.~Houghton}, \bibinfo{author}{K.~Hsu}, \bibinfo{author}{S.~Hu}, \bibinfo{author}{X.~Hu}, \bibinfo{author}{J.~Huizinga}, \bibinfo{author}{S.~Jain}, \bibinfo{author}{S.~Jain}, \bibinfo{author}{J.~Jang}, \bibinfo{author}{A.~Jiang}, \bibinfo{author}{R.~Jiang}, \bibinfo{author}{H.~Jin}, \bibinfo{author}{D.~Jin},
  \bibinfo{author}{S.~Jomoto}, \bibinfo{author}{B.~Jonn}, \bibinfo{author}{H.~Jun}, \bibinfo{author}{T.~Kaftan}, \bibinfo{author}{{\L}.~Kaiser}, \bibinfo{author}{A.~Kamali}, \bibinfo{author}{I.~Kanitscheider}, \bibinfo{author}{N.~S. Keskar}, \bibinfo{author}{T.~Khan}, \bibinfo{author}{L.~Kilpatrick}, \bibinfo{author}{J.~W. Kim}, \bibinfo{author}{C.~Kim}, \bibinfo{author}{Y.~Kim}, \bibinfo{author}{J.~H. Kirchner}, \bibinfo{author}{J.~Kiros}, \bibinfo{author}{M.~Knight}, \bibinfo{author}{D.~Kokotajlo}, \bibinfo{author}{{\L}.~Kondraciuk}, \bibinfo{author}{A.~Kondrich}, \bibinfo{author}{A.~Konstantinidis}, \bibinfo{author}{K.~Kosic}, \bibinfo{author}{G.~Krueger}, \bibinfo{author}{V.~Kuo}, \bibinfo{author}{M.~Lampe}, \bibinfo{author}{I.~Lan}, \bibinfo{author}{T.~Lee}, \bibinfo{author}{J.~Leike}, \bibinfo{author}{J.~Leung}, \bibinfo{author}{D.~Levy}, \bibinfo{author}{C.~M. Li}, \bibinfo{author}{R.~Lim}, \bibinfo{author}{M.~Lin}, \bibinfo{author}{S.~Lin}, \bibinfo{author}{M.~Litwin}, \bibinfo{author}{T.~Lopez},
  \bibinfo{author}{R.~Lowe}, \bibinfo{author}{P.~Lue}, \bibinfo{author}{A.~Makanju}, \bibinfo{author}{K.~Malfacini}, \bibinfo{author}{S.~Manning}, \bibinfo{author}{T.~Markov}, \bibinfo{author}{Y.~Markovski}, \bibinfo{author}{B.~Martin}, \bibinfo{author}{K.~Mayer}, \bibinfo{author}{A.~Mayne}, \bibinfo{author}{B.~McGrew}, \bibinfo{author}{S.~M. McKinney}, \bibinfo{author}{C.~McLeavey}, \bibinfo{author}{P.~McMillan}, \bibinfo{author}{J.~McNeil}, \bibinfo{author}{D.~Medina}, \bibinfo{author}{A.~Mehta}, \bibinfo{author}{J.~Menick}, \bibinfo{author}{L.~Metz}, \bibinfo{author}{A.~Mishchenko}, \bibinfo{author}{P.~Mishkin}, \bibinfo{author}{V.~Monaco}, \bibinfo{author}{E.~Morikawa}, \bibinfo{author}{D.~Mossing}, \bibinfo{author}{T.~Mu}, \bibinfo{author}{M.~Murati}, \bibinfo{author}{O.~Murk}, \bibinfo{author}{D.~M{\'e}ly}, \bibinfo{author}{A.~Nair}, \bibinfo{author}{R.~Nakano}, \bibinfo{author}{R.~Nayak}, \bibinfo{author}{A.~Neelakantan}, \bibinfo{author}{R.~Ngo}, \bibinfo{author}{H.~Noh}, \bibinfo{author}{L.~Ouyang},
  \bibinfo{author}{C.~O'Keefe}, \bibinfo{author}{J.~Pachocki}, \bibinfo{author}{A.~Paino}, \bibinfo{author}{J.~Palermo}, \bibinfo{author}{A.~Pantuliano}, \bibinfo{author}{G.~Parascandolo}, \bibinfo{author}{J.~Parish}, \bibinfo{author}{E.~Parparita}, \bibinfo{author}{A.~Passos}, \bibinfo{author}{M.~Pavlov}, \bibinfo{author}{A.~Peng}, \bibinfo{author}{A.~Perelman}, \bibinfo{author}{F.~d. A.~B. Peres}, \bibinfo{author}{M.~Petrov}, \bibinfo{author}{H.~P. d.~O. Pinto}, \bibinfo{author}{Michael}, \bibinfo{author}{Pokorny}, \bibinfo{author}{M.~Pokrass}, \bibinfo{author}{V.~H. Pong}, \bibinfo{author}{T.~Powell}, \bibinfo{author}{A.~Power}, \bibinfo{author}{B.~Power}, \bibinfo{author}{E.~Proehl}, \bibinfo{author}{R.~Puri}, \bibinfo{author}{A.~Radford}, \bibinfo{author}{J.~Rae}, \bibinfo{author}{A.~Ramesh}, \bibinfo{author}{C.~Raymond}, \bibinfo{author}{F.~Real}, \bibinfo{author}{K.~Rimbach}, \bibinfo{author}{C.~Ross}, \bibinfo{author}{B.~Rotsted}, \bibinfo{author}{H.~Roussez}, \bibinfo{author}{N.~Ryder},
  \bibinfo{author}{M.~Saltarelli}, \bibinfo{author}{T.~Sanders}, \bibinfo{author}{S.~Santurkar}, \bibinfo{author}{G.~Sastry}, \bibinfo{author}{H.~Schmidt}, \bibinfo{author}{D.~Schnurr}, \bibinfo{author}{J.~Schulman}, \bibinfo{author}{D.~Selsam}, \bibinfo{author}{K.~Sheppard}, \bibinfo{author}{T.~Sherbakov}, \bibinfo{author}{J.~Shieh}, \bibinfo{author}{S.~Shoker}, \bibinfo{author}{P.~Shyam}, \bibinfo{author}{S.~Sidor}, \bibinfo{author}{E.~Sigler}, \bibinfo{author}{M.~Simens}, \bibinfo{author}{J.~Sitkin}, \bibinfo{author}{K.~Slama}, \bibinfo{author}{I.~Sohl}, \bibinfo{author}{B.~Sokolowsky}, \bibinfo{author}{Y.~Song}, \bibinfo{author}{N.~Staudacher}, \bibinfo{author}{F.~P. Such}, \bibinfo{author}{N.~Summers}, \bibinfo{author}{I.~Sutskever}, \bibinfo{author}{J.~Tang}, \bibinfo{author}{N.~Tezak}, \bibinfo{author}{M.~B. Thompson}, \bibinfo{author}{P.~Tillet}, \bibinfo{author}{A.~Tootoonchian}, \bibinfo{author}{E.~Tseng}, \bibinfo{author}{P.~Tuggle}, \bibinfo{author}{N.~Turley}, \bibinfo{author}{J.~Tworek},
  \bibinfo{author}{J.~F.~C. Uribe}, \bibinfo{author}{A.~Vallone}, \bibinfo{author}{A.~Vijayvergiya}, \bibinfo{author}{C.~Voss}, \bibinfo{author}{C.~Wainwright}, \bibinfo{author}{J.~J. Wang}, \bibinfo{author}{A.~Wang}, \bibinfo{author}{B.~Wang}, \bibinfo{author}{J.~Ward}, \bibinfo{author}{J.~Wei}, \bibinfo{author}{C.~J. Weinmann}, \bibinfo{author}{A.~Welihinda}, \bibinfo{author}{P.~Welinder}, \bibinfo{author}{J.~Weng}, \bibinfo{author}{L.~Weng}, \bibinfo{author}{M.~Wiethoff}, \bibinfo{author}{D.~Willner}, \bibinfo{author}{C.~Winter}, \bibinfo{author}{S.~Wolrich}, \bibinfo{author}{H.~Wong}, \bibinfo{author}{L.~Workman}, \bibinfo{author}{S.~Wu}, \bibinfo{author}{J.~Wu}, \bibinfo{author}{M.~Wu}, \bibinfo{author}{K.~Xiao}, \bibinfo{author}{T.~Xu}, \bibinfo{author}{S.~Yoo}, \bibinfo{author}{K.~Yu}, \bibinfo{author}{Q.~Yuan}, \bibinfo{author}{W.~Zaremba}, \bibinfo{author}{R.~Zellers}, \bibinfo{author}{C.~Zhang}, \bibinfo{author}{M.~Zhang}, \bibinfo{author}{S.~Zhao}, \bibinfo{author}{T.~Zheng},
  \bibinfo{author}{J.~Zhuang}, \bibinfo{author}{W.~Zhuk}, \bibinfo{author}{B.~Zoph}, \bibinfo{title}{{{GPT-4 Technical Report}}}, \bibinfo{year}{2024}. \DOIprefix\doi{10.48550/arXiv.2303.08774}. \href{http://arxiv.org/abs/2303.08774}{{\tt arXiv:2303.08774}}.
\bibitem[{Touvron et~al.(2023)Touvron, Lavril, Izacard, Martinet, Lachaux, Lacroix, Rozi{\`e}re, Goyal, Hambro, Azhar, Rodriguez, Joulin, Grave, and Lample}]{touvron2023llama}
\bibinfo{author}{H.~Touvron}, \bibinfo{author}{T.~Lavril}, \bibinfo{author}{G.~Izacard}, \bibinfo{author}{X.~Martinet}, \bibinfo{author}{M.-A. Lachaux}, \bibinfo{author}{T.~Lacroix}, \bibinfo{author}{B.~Rozi{\`e}re}, \bibinfo{author}{N.~Goyal}, \bibinfo{author}{E.~Hambro}, \bibinfo{author}{F.~Azhar}, \bibinfo{author}{A.~Rodriguez}, \bibinfo{author}{A.~Joulin}, \bibinfo{author}{E.~Grave}, \bibinfo{author}{G.~Lample}, \bibinfo{title}{{{LLaMA}}: {{Open}} and {{Efficient Foundation Language Models}}}, \bibinfo{year}{2023}. \DOIprefix\doi{10.48550/arXiv.2302.13971}. \href{http://arxiv.org/abs/2302.13971}{{\tt arXiv:2302.13971}}.
\bibitem[{Elsafoury and Katsigiannis(2024)}]{elsafoury2024bias}
\bibinfo{author}{F.~Elsafoury}, \bibinfo{author}{S.~Katsigiannis}, \bibinfo{title}{On {{Bias}} and {{Fairness}} in {{NLP}}: {{Investigating}} the {{Impact}} of {{Bias}} and {{Debiasing}} in {{Language Models}} on the {{Fairness}} of {{Toxicity Detection}}}, \bibinfo{year}{2024}. \DOIprefix\doi{10.48550/arXiv.2305.12829}. \href{http://arxiv.org/abs/2305.12829}{{\tt arXiv:2305.12829}}.
\bibitem[{Bender et~al.(2021)Bender, Gebru, {McMillan-Major}, and Shmitchell}]{bender2021dangers}
\bibinfo{author}{E.~M. Bender}, \bibinfo{author}{T.~Gebru}, \bibinfo{author}{A.~{McMillan-Major}}, \bibinfo{author}{S.~Shmitchell},
\newblock \bibinfo{title}{On the {{Dangers}} of {{Stochastic Parrots}}: {{Can Language Models Be Too Big}}?},
\newblock in: \bibinfo{booktitle}{Proceedings of the 2021 {{ACM Conference}} on {{Fairness}}, {{Accountability}}, and {{Transparency}}}, {{FAccT}} '21, \bibinfo{publisher}{Association for Computing Machinery}, \bibinfo{address}{New York, NY, USA}, \bibinfo{year}{2021}, pp. \bibinfo{pages}{610--623}. \DOIprefix\doi{10.1145/3442188.3445922}.
\bibitem[{ris(2025)}]{rise}
\bibinfo{title}{Rise of the {{Newsbots}}: {{AI-Generated News Websites Proliferating Online}}}, \bibinfo{year}{2025}.
\bibitem[{jun(2025)}]{junk}
\bibinfo{title}{Junk websites filled with {{AI-generated}} text are pulling in money from programmatic ads}, \bibinfo{howpublished}{https://www.technologyreview.com/2023/06/26/1075504/junk-websites-filled-with-ai-generated-text-are-pulling-in-money-from-programmatic-ads/}, \bibinfo{year}{2025}.
\bibitem[{Weidinger et~al.(2021)Weidinger, Mellor, Rauh, Griffin, Uesato, Huang, Cheng, Glaese, Balle, Kasirzadeh, Kenton, Brown, Hawkins, Stepleton, Biles, Birhane, Haas, Rimell, Hendricks, Isaac, Legassick, Irving, and Gabriel}]{weidinger2021ethical}
\bibinfo{author}{L.~Weidinger}, \bibinfo{author}{J.~Mellor}, \bibinfo{author}{M.~Rauh}, \bibinfo{author}{C.~Griffin}, \bibinfo{author}{J.~Uesato}, \bibinfo{author}{P.-S. Huang}, \bibinfo{author}{M.~Cheng}, \bibinfo{author}{M.~Glaese}, \bibinfo{author}{B.~Balle}, \bibinfo{author}{A.~Kasirzadeh}, \bibinfo{author}{Z.~Kenton}, \bibinfo{author}{S.~Brown}, \bibinfo{author}{W.~Hawkins}, \bibinfo{author}{T.~Stepleton}, \bibinfo{author}{C.~Biles}, \bibinfo{author}{A.~Birhane}, \bibinfo{author}{J.~Haas}, \bibinfo{author}{L.~Rimell}, \bibinfo{author}{L.~A. Hendricks}, \bibinfo{author}{W.~Isaac}, \bibinfo{author}{S.~Legassick}, \bibinfo{author}{G.~Irving}, \bibinfo{author}{I.~Gabriel}, \bibinfo{title}{Ethical and social risks of harm from {{Language Models}}}, \bibinfo{year}{2021}. \DOIprefix\doi{10.48550/arXiv.2112.04359}. \href{http://arxiv.org/abs/2112.04359}{{\tt arXiv:2112.04359}}.
\bibitem[{Sadasivan et~al.(2024)Sadasivan, Kumar, Balasubramanian, Wang, and Feizi}]{sadasivan2024can}
\bibinfo{author}{V.~S. Sadasivan}, \bibinfo{author}{A.~Kumar}, \bibinfo{author}{S.~Balasubramanian}, \bibinfo{author}{W.~Wang}, \bibinfo{author}{S.~Feizi}, \bibinfo{title}{Can {{AI-Generated Text}} be {{Reliably Detected}}?}, \bibinfo{year}{2024}. \DOIprefix\doi{10.48550/arXiv.2303.11156}. \href{http://arxiv.org/abs/2303.11156}{{\tt arXiv:2303.11156}}.
\bibitem[{Kumarage et~al.(2024)Kumarage, Agrawal, Sheth, Moraffah, Chadha, Garland, and Liu}]{kumarage2024survey}
\bibinfo{author}{T.~Kumarage}, \bibinfo{author}{G.~Agrawal}, \bibinfo{author}{P.~Sheth}, \bibinfo{author}{R.~Moraffah}, \bibinfo{author}{A.~Chadha}, \bibinfo{author}{J.~Garland}, \bibinfo{author}{H.~Liu}, \bibinfo{title}{A {{Survey}} of {{AI-generated Text Forensic Systems}}: {{Detection}}, {{Attribution}}, and {{Characterization}}}, \bibinfo{year}{2024}. \href{http://arxiv.org/abs/2403.01152}{{\tt arXiv:2403.01152}}.
\bibitem[{Mao et~al.(2023)Mao, Vondrick, Wang, and Yang}]{mao2023raidar}
\bibinfo{author}{C.~Mao}, \bibinfo{author}{C.~Vondrick}, \bibinfo{author}{H.~Wang}, \bibinfo{author}{J.~Yang},
\newblock \bibinfo{title}{Raidar: {{geneRative AI Detection viA Rewriting}}},
\newblock in: \bibinfo{booktitle}{The {{Twelfth International Conference}} on {{Learning Representations}}}, \bibinfo{year}{2023}.
\bibitem[{Horne et~al.(2018)Horne, Dron, Khedr, and Adali}]{horne2018assessing}
\bibinfo{author}{B.~D. Horne}, \bibinfo{author}{W.~Dron}, \bibinfo{author}{S.~Khedr}, \bibinfo{author}{S.~Adali},
\newblock \bibinfo{title}{Assessing the {{News Landscape}}: {{A Multi-Module Toolkit}} for {{Evaluating}} the {{Credibility}} of {{News}}},
\newblock in: \bibinfo{booktitle}{Companion {{Proceedings}} of the {{The Web Conference}} 2018}, {{WWW}} '18, \bibinfo{publisher}{International World Wide Web Conferences Steering Committee}, \bibinfo{address}{Republic and Canton of Geneva, CHE}, \bibinfo{year}{2018}, pp. \bibinfo{pages}{235--238}. \DOIprefix\doi{10.1145/3184558.3186987}.
\bibitem[{Opara(2024)}]{opara2024styloai}
\bibinfo{author}{C.~Opara},
\newblock \bibinfo{title}{Styloai: Distinguishing ai-generated content with stylometric analysis},
\newblock in: \bibinfo{booktitle}{International conference on artificial intelligence in education}, \bibinfo{organization}{Springer}, \bibinfo{year}{2024}, pp. \bibinfo{pages}{105--114}.
\bibitem[{Horne and Adali(2017)}]{horne2017just}
\bibinfo{author}{B.~Horne}, \bibinfo{author}{S.~Adali},
\newblock \bibinfo{title}{This just in: {{Fake}} news packs a lot in title, uses simpler, repetitive content in text body, more similar to satire than real news},
\newblock in: \bibinfo{booktitle}{Proceedings of the International {{AAAI}} Conference on Web and Social Media}, volume~\bibinfo{volume}{11}, \bibinfo{year}{2017}, pp. \bibinfo{pages}{759--766}.
\bibitem[{Tausczik and Pennebaker(2010)}]{tausczik2010psychological}
\bibinfo{author}{Y.~R. Tausczik}, \bibinfo{author}{J.~W. Pennebaker},
\newblock \bibinfo{title}{The psychological meaning of words: {{LIWC}} and computerized text analysis methods},
\newblock \bibinfo{journal}{Journal of Language and Social Psychology} \bibinfo{volume}{29} (\bibinfo{year}{2010}) \bibinfo{pages}{24--54}. \DOIprefix\doi{10.1177/0261927X09351676}.
\bibitem[{Thelwall et~al.(2010)Thelwall, Buckley, Paltoglou, Cai, and Kappas}]{thelwall2010sentiment}
\bibinfo{author}{M.~Thelwall}, \bibinfo{author}{K.~Buckley}, \bibinfo{author}{G.~Paltoglou}, \bibinfo{author}{D.~Cai}, \bibinfo{author}{A.~Kappas},
\newblock \bibinfo{title}{Sentiment strength detection in short informal text},
\newblock \bibinfo{journal}{Journal of the American society for information science and technology} \bibinfo{volume}{61} (\bibinfo{year}{2010}) \bibinfo{pages}{2544--2558}.
\bibitem[{Roy et~al.(2025{\natexlab{a}})Roy, Singh, Aziz, Bajpai, Imanpour, Biswas, Wanaskar, Patwa, Ghosh, Dixit, Pal, Rawte, Garimella, Das, Sheth, Sharma, Reganti, Jain, and Chadha}]{roy2025defactify_overview_text}
\bibinfo{author}{R.~Roy}, \bibinfo{author}{G.~Singh}, \bibinfo{author}{A.~Aziz}, \bibinfo{author}{S.~Bajpai}, \bibinfo{author}{N.~Imanpour}, \bibinfo{author}{S.~Biswas}, \bibinfo{author}{K.~Wanaskar}, \bibinfo{author}{P.~Patwa}, \bibinfo{author}{S.~Ghosh}, \bibinfo{author}{S.~Dixit}, \bibinfo{author}{N.~R. Pal}, \bibinfo{author}{V.~Rawte}, \bibinfo{author}{R.~Garimella}, \bibinfo{author}{A.~Das}, \bibinfo{author}{A.~Sheth}, \bibinfo{author}{V.~Sharma}, \bibinfo{author}{A.~N. Reganti}, \bibinfo{author}{V.~Jain}, \bibinfo{author}{A.~Chadha},
\newblock \bibinfo{title}{Overview of text counter turing test: Ai generated text detection},
\newblock in: \bibinfo{booktitle}{proceedings of {D}e{F}actify 4: Fourth workshop on Multimodal Fact-Checking and Hate Speech Detection}, \bibinfo{publisher}{CEUR}, \bibinfo{year}{2025}{\natexlab{a}}.
\bibitem[{Roy et~al.(2025{\natexlab{b}})Roy, Singh, Aziz, Bajpai, Imanpour, Biswas, Wanaskar, Patwa, Ghosh, Dixit, Pal, Rawte, Garimella, Das, Sheth, Sharma, Reganti, Jain, and Chadha}]{roy2025defactify_dataset_text}
\bibinfo{author}{R.~Roy}, \bibinfo{author}{G.~Singh}, \bibinfo{author}{A.~Aziz}, \bibinfo{author}{S.~Bajpai}, \bibinfo{author}{N.~Imanpour}, \bibinfo{author}{S.~Biswas}, \bibinfo{author}{K.~Wanaskar}, \bibinfo{author}{P.~Patwa}, \bibinfo{author}{S.~Ghosh}, \bibinfo{author}{S.~Dixit}, \bibinfo{author}{N.~R. Pal}, \bibinfo{author}{V.~Rawte}, \bibinfo{author}{R.~Garimella}, \bibinfo{author}{A.~Das}, \bibinfo{author}{A.~Sheth}, \bibinfo{author}{V.~Sharma}, \bibinfo{author}{A.~N. Reganti}, \bibinfo{author}{V.~Jain}, \bibinfo{author}{A.~Chadha},
\newblock \bibinfo{title}{Defactify-text: A comprehensive dataset for human vs. ai generated text detection},
\newblock in: \bibinfo{booktitle}{proceedings of {D}e{F}actify 4: Fourth workshop on Multimodal Fact-Checking and Hate Speech Detection}, \bibinfo{publisher}{CEUR}, \bibinfo{year}{2025}{\natexlab{b}}.
\bibitem[{Oktafia and Nugroho(2022)}]{oktafia2022comparison}
\bibinfo{author}{O.~Oktafia}, \bibinfo{author}{R.~S.~A. Nugroho},
\newblock \bibinfo{title}{{{COMPARISON OF SUPPORT VECTOR MACHINE}}({{SVM}}), {{XGBOOST AND RANDOM FOREST FOR SENTIMENT ANALYSIS OF BUMBLE APP USER COMMENTS}}},
\newblock \bibinfo{journal}{Proxies : Jurnal Informatika} \bibinfo{volume}{6} (\bibinfo{year}{2022}) \bibinfo{pages}{32--46}. \DOIprefix\doi{10.24167/proxies.v6i1.12453}.
\bibitem[{Umam et~al.(2024)Umam, Handoko, and Isinkaye}]{umam2024performance}
\bibinfo{author}{C.~Umam}, \bibinfo{author}{L.~B. Handoko}, \bibinfo{author}{F.~O. Isinkaye},
\newblock \bibinfo{title}{Performance {{Analysis}} of {{Support Vector Classification}} and {{Random Forest}} in {{Phishing Email Classification}}},
\newblock \bibinfo{journal}{Scientific Journal of Informatics} \bibinfo{volume}{11} (\bibinfo{year}{2024}) \bibinfo{pages}{367--374}. \DOIprefix\doi{10.15294/sji.v11i2.3301}.

\end{thebibliography}

\end{document}